\documentclass[lettersize, journal]{IEEEtran}
\usepackage{amsmath,amsfonts}
\usepackage{algorithmic}
\usepackage{algorithm}
\usepackage{array}
\usepackage[caption=false,font=normalsize,labelfont=sf,textfont=sf]{subfig}
\usepackage{textcomp}
\usepackage{stfloats}
\usepackage{url}
\usepackage{verbatim}
\usepackage{graphicx}
\usepackage{cite}

\usepackage[utf8]{inputenc}
\usepackage{hyperref}
\usepackage{float}
\usepackage{xcolor}
\usepackage{soul}
\usepackage{tabulary}
\usepackage{tablefootnote}

\newtheorem{definition}{Definition}[section]

\begin{document}

\title{Detection of Abuse in Financial Transaction Descriptions Using Machine Learning}

\author{Anna Leontjeva, Genevieve Richards, Kaavya Sriskandaraja, Jessica Perchman, Luiz Pizzato \thanks{All authors are with Artificial Intelligence Labs, Commonwealth Bank of Australia, Sydney, Australia}}



\maketitle

\begin{abstract}
Since introducing changes to the New Payments Platform (NPP) to include longer messages as payment descriptions, it has been identified that people are now using it for communication, and in some cases, the system was being used as a targeted form of domestic and family violence. This type of tech-assisted abuse poses new challenges in terms of identification, actions and approaches to rectify this behaviour. Commonwealth Bank of Australia's Artificial Intelligence Labs team (CBA AI Labs) has developed a new system using advances in deep learning models for natural language processing (NLP) to create a powerful abuse detector that periodically scores all the transactions, and identifies cases of high-risk abuse in millions of records. In this paper, we describe the problem of tech-assisted abuse in the context of banking services, outline the developed model and its performance, and the operating framework more broadly. 
\end{abstract}


\section{Introduction}

\subsection{Technology Assisted Abuse}
\IEEEPARstart{D}{igital} communication plays an increasingly important role in everyday life. As of 2021, 4.55 billion people are active social media users, equating to 57.6\% of the world population~\cite{social}. The prevalence and variety of digital communication have given us the ability to contact someone 24 hours a day through many different ways such as social media, text-messaging, and email. Although this has increased convenience for a lot of people, it has also presented significant challenges for personal security and privacy, and in particular for domestic violence victims/survivors \cite{dragiewicz2019domestic}.

Technology-facilitated abuse is commonly defined as the use of technology such as mobile, online or other digital technologies, as a tool for people to engage in behaviours such as coercive control, intimidation, stalking, monitoring, psychological and emotional abuse, consistent harassment and unwanted contact, sexual harassment, to cause harm and distress to the recipient \cite{fiolet2021exploring}. This term can be extended to include broader forms of online harassment and cyber bullying; however, it is typically focused on gendered violence (domestic violence) \cite{flynn2021technology}. The impacts of technology-facilitated abuse on the recipient can include depression, worthlessness, fatigue, self-harm, traumatisation, fear, isolation, emotional distress and more. There are also reported economic impacts, functional harms and an intrusion on the recipient's personal freedom \cite{flynn2021technology}.

\subsection{Technology assisted abuse in Banking}
Modern payment systems have increased the speed of financial transactions and also enabled richer descriptions of those transactions \cite{austrac2021}. The introduction of the New Payment Platform (NPP) in Australia in 2018\footnote{See: \url{https://nppa.com.au/the-platform/}} allows a person or a business to conduct a transfer to others in real-time and include up to 280 UTF-8 characters for payment description and an additional 35 printable ASCI characters for payment reference. NPP has also provided customers with the ability to set up a simple identifier (PayID®) for their accounts, such as a mobile number or an email address, so that they no longer need to remember their Bank State Number (BSB)\footnote{BSB is a number that indicates the bank and branch that holds ones account in Australia, facilitating a transaction between banks.\\\ \\} and account number. It resulted in a simple and fast way for people to transfer funds to each other in Australia.  As of June 2022, 107 Australian financial institutions use these services, and more than 10 million PayIDs have been registered by customers and businesses \cite{reservebank2022payments}. New technologies such as PayID simplified banking increasing the ease and volume of transactions, however, it also provided perpetrators another tool to use for abuse. 

In early 2020, we as the Commonwealth Bank of Australia (CBA) identified the use of real-time transactions as a means of communication between individuals, typically through the use of low value transactions. We found that more than 8,000 customers in a three-month period had received multiple low-value deposits with messages in the transaction description that were potentially abusive. We identified that the intent of the messages ranged from ``jokes'' using profanity to serious threats or references to domestic abuse and family violence \cite{austrac2021}. Utilizing transaction descriptions as a mode of either criminal communication or abuse rather than as means to transfer funds is being detected in financial institutions across the world. 
Australian Transaction Reports and Analysis Centre (AUSTRAC) Fintel Alliance report \cite{austrac2021} notes that it is not unique to the Australian banking industry.  For example, several Brazillian news groups report the arrest of a man harassing a young woman through bank transfers after having his number blocked \cite{gabriel_news}. We can see that any payment that contains a free text field to be completed by the sender and viewed by the recipient can be a vehicle for criminal communication.

\subsection{Role of the Bank}
Although online banking was never intended to be used as a digital communication technology, its occurrence has meant that financial institutions had to take action in order to protect those being abused. Initial responses by financial institutions have involved actions such updating their terms and conditions to include references to abusive transactions and introducing real-time word blocks from reference lists~\cite{majorbanks}. These measures have shown to significantly reduce the use of profanity and some abuse in transaction description, however, they have not completely stop serious abuse from happening.   

Although these solutions have seen a reduction in profanity used in online banking transactions, they are not stopping abusers with the intention to cause harm or distress to the recipients as they have simply learnt how to circumvent these initial solutions. For example, the word \emph{unblock}, which is associated with these abusive payments, was observed to be modified to \emph{un-block}, \emph{u.n.b.l.o.c.k} and other versions to bypass it. Because of this, we decided to protect our customers by building a monitoring system that can work in the background identifying cases of serious abuse that may need to be further investigated.

Building a system that identifies abuse involves, among many things, the definition of abuse, and the design of a system and processes that can help the victims and dissuade the perpetrator. To proactively stop abusers or to reach out and provide support to the affected customers, we first need to identify these cases. There is a lot of complexity in this task alone including the volume of transactions sent each day, understanding the context of the transaction sent and the nuances of the language and behaviour used. We have addressed the issue by using a multi-step approach. First, the model is applied to score all the transactions. The cases with the highest score are then sent to a dedicated team of customer vulnerability specialists that manually review and contact the victims of abuse identified by the model. The team will then take the most appropriate action, for example, it may involve contacting the victim, as well as sending warning letters to the perpetrator and let them know their behaviour is not tolerated. In some cases it might involve welfare checks to ensure their safety and to gain their consent to take further action. Due to the capacity and complexity of cases and interventions offered, the team is only able to manually review and process a limited number of cases a month. Therefore, it is crucial that we control the number of false positive cases ensuring we detect all the true positive cases.

Due to the novelty of the problem, the current approach doesn't contain comparison with the other models, and we believe can be improved by efforts of the wider research community. However, the current work establishes a solid baseline to compare it to. We also hope that it helps to adopt these techniques in the other financial institutions that are currently utilising simple filters and keyword detection that can be easily overcome and bypassed.

\section{Problem Statement}
\label{section:problem}

In this paper we propose an approach to the problem of high-risk abuse case detection in the banking payment systems using a combination of features from different deep learning models. Despite the fact that technology assisted abuse is not a new problem and some research has been conducted to investigate it (see Section~\ref{section:literature} for more details), this type of abuse using banking transactions was only recently identified and poses a new set of challenges. One of the biggest challenges is the sensitivity of the matter. It should be handled with uttermost care considering that both action and no-action can be potentially dangerous. Bank transactions are different to social media messages that can be easily deleted or blocked. The transactions have much longer ``life-span''. They might be visible to someone beyond the victim.  They might be delivered in printed form.  They might be used as evidence for people's applications to loans and other services, which can cause re-traumatisation by revisiting them.
Similarly, the victim might have much lower tolerance towards the abuser's behaviour and all these situations can be difficult to deal with. 

In terms of actions taken, the bank often contacts abusers asking them to stop. If the behaviour continues, it is possible to \emph{unbank} a customer. However, differently to social media bans, unbanking is a decision of a financial institution to ends its relationship with a particular individual and could lead to serious consequences affecting peoples' lives. To complicate things further, transaction descriptions are often limiting in context and often open to the interpretations. Therefore, a dedicated team has to investigate and approach to each case individually, which is a labour-intense process that leads to the prioritisation of the cases. This is known as a human-in-the-loop system, defined by needing both human and machine performance to contribute to improving the overall system results \cite{humanloop}. Therefore, in this paper we focus on the framework of detecting high-risk cases that need to be prioritised, allowing us to adhere to Australia's AI Ethics Principles of reliability and safety \cite{AIpolicy}.

\begin{definition}
\label{definition:highrisk}
	High-risk cases of abuse are defined by the severity and volume of the following:
	\begin{itemize}
		\item the presence of repetitive, abusive, degrading or hateful comments about a person or persons
		\item threats of physical or sexual violence to a person
		\item threats of self-harm
		\item endangering or causing distress to a minor
		\item repeated or unwanted sexual requests to a person.
	\end{itemize}
\end{definition}

\section{Literature review}
\label{section:literature}

A report by \cite{penzeymoog2021technology} provides an in-depth investigation of different technology utilized by abusers to commit technology facilitated abuse related to Domestic Violence. The report explores types of abuse associated with coercive control, financial abuse, smart homes and stalking, and how these are misused by abusers. The report also provides a framework for inclusive safety when designing technology systems however does not suggest any solutions around how to identify when a system is being misused. Although the report touches on how financial abuse can happen in banking systems it does not explore a problem of transactions descriptions being utilized by abusers to send abusive messages and exhibit control and stalking behaviours. 
 
 The problem of detecting abusive messages in bank transaction descriptions is novel. While similar problems of detecting offensive language, toxicity levels in text, bullying and hate speech has been a subject of research over the past 20 years, it has mostly been in the context of social network moderation, for example, employing machine learning techniques to identify this type of content from Twitter \cite{zampieri2019semeval} and Facebook  \cite{wang2021entailment}.

In a similar fashion to the issue of abusive messages, data within social networks is highly unstructured, informal, and often misspelled, therefore, papers such as~\cite{chen2012detecting} have utilised natural language processing techniques to detect both lexical and syntactic features of sentences. Branching out from solely using features from the text, \cite{chen2012detecting} used style, structure and posting pattern features to improve detection of offensive messages. \cite{chatzakou2017measuring} used joy, emoticons, uppercase, number of followers, amongst other features.
 
\cite{wang2021entailment} outlines a new approach called Entailment as Few Shot Learner (EFL). With the aim to improve language models as few-shot learners, the approach involves converting class labels into a natural language sentence which is used to describe the label, and determine whether the label entails the description. The EFL approach can also leverage techniques such as unsupervised contrastive data augmentation and can be extended to multilingual few-shot learning. \cite{raj2021cyberbullying} proposes a novel shallow neural network using GloVe embeddings on Wikipedia public datasets to classify whether the comments are toxic or are instances of attack in cyber bullying context. 

\cite{zampieri2019semeval} leveraged machine learning to detect targeted vs untargeted offensive language. This was done by creating a three-level annotation schema, corresponding to three subtasks. The first Subtask A focussed on purely the language in a dataset, classifying it as either offensive or non offensive. Subtask B further classified the data as targeted or untargeted, i.e. general offensive language or hate speech, and Subtask C classifies whether the hate speech was targeted at an individual or a group.

Other techniques to detect abuse have leveraged systems based on pre-trained language models such as RoBERTa and BERT \cite{devlin2018bert}, which have reached new state-of-the-art performances on numerous tasks \cite{liu2019roberta}. In \cite{bodapati2019neural}, a BERT model fine-tuned with binary cross-entropy loss was used to identify abusive language in Twitter Hatespeech and Wikipedia datasets. BERT embedded models outperformed other embeddings such as fastText, TextCNN and TextCNN + Character n-grams. One issue that was found with pre-trained models is that they are trained on general datasets, so they have limitations on domain-specific language tasks. Re-training pre-trained models on domain-specific datasets is a popular method to address this as seen in \cite{azzouza2019twitterbert}. This is especially useful contexts such as abusive language detection where there is not enough data to train a BERT-like model from scratch. Their model, `TweetBERT', was re-trained on a Twitter-based corpus and outperformed other BERT based models when analysing Twitter content. 

Paper~\cite{csiro2020cyberbullying} outlined another method to improve BERT based models in order to detect instances of cyberbullying and harmful speech on Australian-based Twitter data. This was done through appending additional features onto BERT as special tokens. The features included emoji paths, metadata such as user information (e.g. age, gender, number of posts), data on their network (e.g. number of follower and friends) and their power (followers/friends ratio). Results showed that BERT with the extra tokens (BERT + emoji + network + power) yielded the most accuracy. 

We observed that prior work focuses on identifying instances of abusive messaging instead of the abusive relationship. As mentioned in Definition~\ref{definition:highrisk} of High-Risk abuse these transactions descriptions need to be relatively consistent and occur in a higher frequency. In a similar fashion, some papers have implemented techniques on user-centric data to identify potentially abusive users, rather than lone instances of abusive language. For example, \cite{ribeiro2017like} used graph machine learning to identify hateful users. Hateful accounts were characterised using attributes such as creation date, user activity, network centrality, sentiment and lexical analysis, amongst other attributes. The methodology involved using a process based on DeGroot's learning model \cite{ribeiro2017like} to sample users in a neighbourhood, and label them as hateful or non-hateful. There were significant patterns found to be associated with `hateful users', including increased activity and increased frequency in using particular language.

While there is a lot of work that focus on online social networks, there has been no research into detecting abuse in transaction descriptions in a context of financial services. In this work, we leverage several machine learning techniques to not only identify abusive language in transaction descriptions, but identify the transaction relationships of the customers who are using it. 

\section{Data}
\label{section:data}
In this section, we introduce the specifics of the dataset we use as well as the data preprocessing methods. This paper relies on Commonwealth Bank transaction data. We extracted details from the bank's database, including transaction descriptions, the corresponding dollar amounts, date of the transaction, sender and recipient account numbers.
We gathered transaction data from both the new payment platform (NPP) and the non-NPP processes.
This data was used to generate features for the model training, which were aggregated by \emph{relationships}, as described in Section~\ref{section:methodology}. Note, a relationship in this context means a sender and a recipient pair of a transaction; that is, if sender $a$ sends a transaction to recipient $b$, we have the $\langle{a,b}\rangle$ relationship, if $b$ sends a transaction to $a$ this creates a different and new $\langle{b,a}\rangle$ relationship. The number of transactions we used by relationship is defined by the historical time-window we used. In this study we fixed time-window to be one month. 

Our dataset contains 1,039 relationships that were labelled as either (1) highly abusive or (0) non-abusive. Among those unique relationships, 283 were branded as 'highly-abusive' by several domestic violence experts who used the definition of high-risk abuse as a guide (see Definition~\ref{definition:highrisk}). They had an agreement score of 87\%. Negative sampling was created by randomly choosing non-abusive relationships as well as a sample of cases where transactions had ``conversational'' descriptions that do not meet the abuse requirements but were significantly different to normal transactions. Some cases, for example, included customers sending song lyrics to one another or a perfectly natural chat. This was done to avoid using a machine learning model to detect only long messages rather than high-risk abuse. This training set contains data from July 2021 to January 2022. We used this dataset for our experiments and validated our proposed system using k-fold cross validation. It is important to note that there are no overlapping relationship pairs between folds.  

In addition, for an out-of-sample dataset, we extracted one month of transaction data. We used data from the month of February 2022 for this. This demonstrates a model scoring use-case as part of the current business process. In any given month, less than 0.0005\% of cases are abusive, resulting in a highly imbalanced situation. As manually scoring all of these monthly transactions is impractical, the out-of-sample test set was created by labelling the top 50 highest scored relationships of the corresponding month for each of the candidate models, 35 of which turned out to be highly abusive.

\section{Methodology}
\label{section:methodology}
\begin{figure}
	\centering
	\includegraphics[width=1\linewidth]{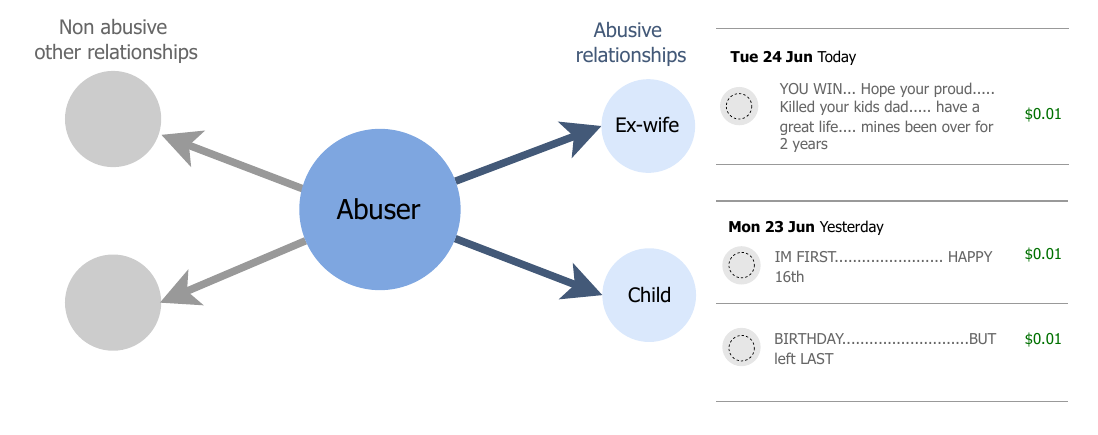}
	\caption{Some abusers intimidate several people. In this case, Abuser's communication with ex-wife and Abuser's communication with his child are considered as separate relationship in our dataset. \bf{Note: the transaction descriptions in this image are not real and were made up as illustrations of how abusive these messages are}. }
	\label{fig:abuser1}	
\end{figure}

In this section we described our approach in more details. The first task was to decide whether it was better to detect abuse at the transaction, customer, or relationship level. The transaction level lacks sufficient textual information to capture the context. Consider a transaction text that says ``I love you''. Without more transactions to observe the dynamics, it's unclear whether this is a case of harassment or a regular message between a couple. However, if this type of description is sent every 5 minutes and the other party requests to stop, the case becomes much clearer. However, if we consider collecting all transactions at the customer level, the abusive information may be diluted. For example, abusive customers frequently harass only one person among their recipients despite having a large network of regular recipients. As a result, we detect abuse at the relationship level, using descriptions gleaned from transactions between each sender-recipient pair. Figure \ref{fig:abuser1} shows an example of an abuser having multiple victims, which in this case we would flag as two distinct relationships of high risk.

\begin{figure}
	\centering
	\includegraphics[width=1\linewidth]{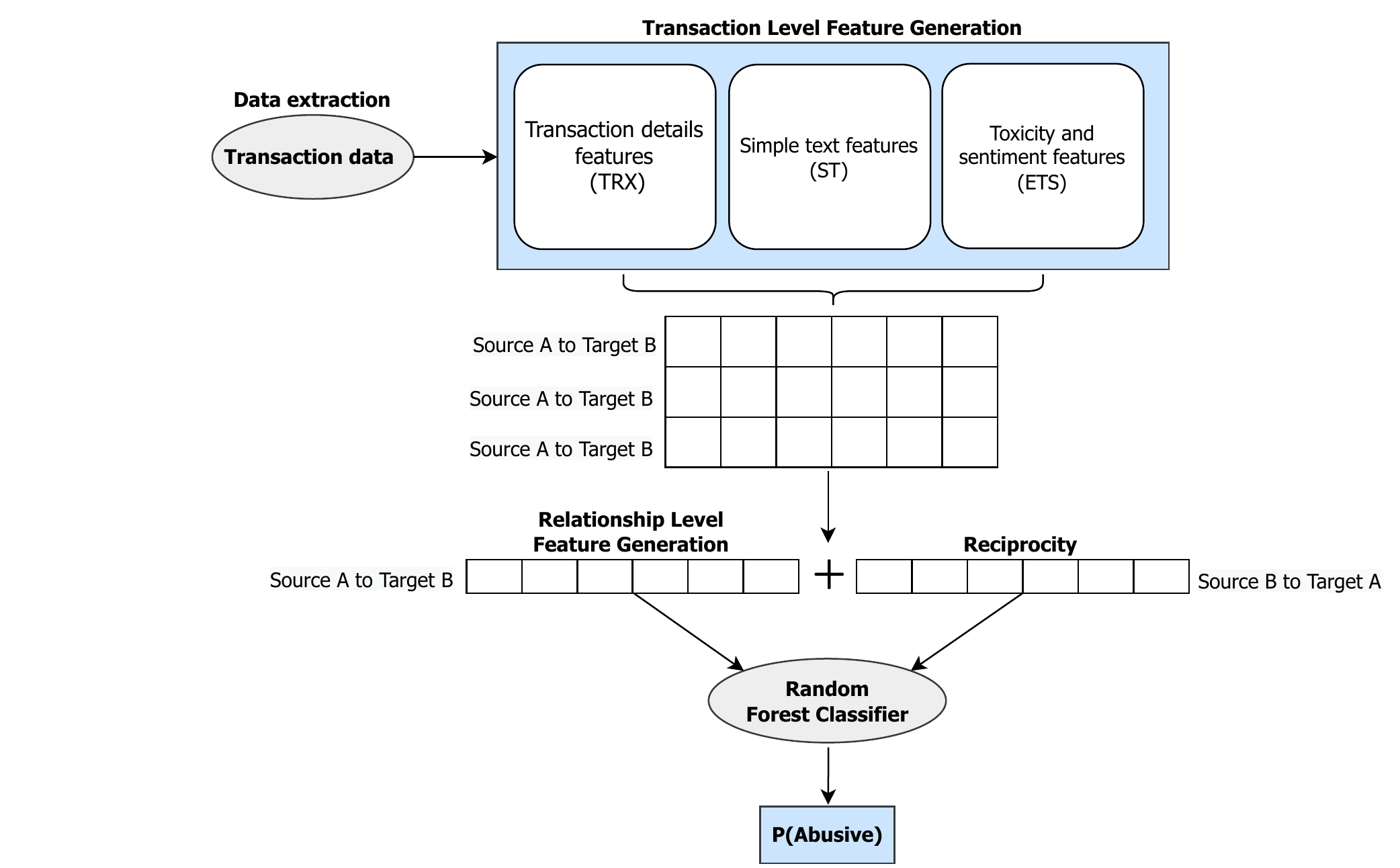}
	\caption{An overview of the system architecture. Three sets of the features are created from the raw data and combined on the level of relationships. The final model takes both original features and reciprocity features.}
	\label{fig:methodology_diagram}	
\end{figure}

Following that, we describe the overall approach developed to detect abuse in transaction descriptions (AITD). It should be noted here that our target is to detect the highly abusive cases. Figure~\ref{fig:methodology_diagram} diagram depicts an overview of our system involving the following steps: 

\begin{enumerate}
	\item Transaction-level feature generation: creating appropriate features from each single transaction description (Section~\ref{section:tlfg})
	\item Relationship-level feature generation: aggregating these features on each relationship (sender-recipient pair) in order to detect abusive customers, not just individual abusive transactions (Section~\ref{section:srplfg})
	\item Incorporating reciprocity information: generating features related to the replies a potential victim might have sent (Section~\ref{section:Reciprocity})
	\item Training a machine learning model to predict the labels: Random Forest model was used to classify relationships as either highly abusive or non-abusive %
\end{enumerate}

\subsection{Transaction-level feature generation} 
\label{section:tlfg}
As described previously, the first step is to generate features on a transaction level. This transaction-level feature vector is then aggregated to create features on a relationship level. There are three types of features involved in our model, as follows:

\begin{itemize}
\item \textbf{Transaction details features (TRX)} are solely related to the specifics of the transaction between sender and receiver, such as: dollar amount transacted, date of the transaction, number of transactions a day, maximum number of transactions per day and the time between maximum and minimum number of transactions. 
\item \textbf{Simple text features (ST)} are related to the basic information we can extract from the transaction description, such as length of the transaction description, upper/lower/mixed case flags, number of words, length of the longest word in the transaction description, does the message contains special characters/numbers, empty description flag, various punctuation and number-related flags.
\item \textbf{Emotion, Toxicity and Sentiment features (ETS)} are features based on three pre-trained language models that are able to provide valuable information for the abuse detection based on the text in the message descriptions. 
Seven \emph{toxicity} features were generated per transaction using the unbiased version of the BERT-based, pre-trained language model Detoxify ~\cite{detoxify}. The unbiased version of Detoxify recognises toxicity and minimises unintended bias of identities. It scores the text on seven categories of toxicity, such as toxicity, severe toxicity, obscene, threat, insult, identity attack and sexual explicit. These scores were then used as the seven toxicity features for the proposed AITD model.
For \emph{emotion} features we used DistilBERT models trained on four data sources, including dailydialog, emotion-stimulus, isear and huggingface emotion datasets~\cite{hugging_emotions}. This pre-trained model determines whether the given text is neutral, joyful, sad, angry, contains love, fear, or surprise emotions. We predicted the scores for each emotion class for each transaction description.
We used VADER (Valence Aware Dictionary for Sentiment Reasoning) for finding the \emph{sentiment} of a transaction description. It indicates both the polarity (positive/negative) and the intensity (strength) of emotion. The sentimental analysis of VADER is based on a dictionary that maps lexical features to emotion intensities known as sentiment scores. A text's sentiment score can be calculated by adding the intensity of each word in the text. VADER's Sentiment intensity analyser accepts a string and returns a dictionary of scores in four categories, including positive, negative, compound, and neutral ~\cite{hutto2014vader}. 
\end{itemize} 

\subsection{Relationship level feature generation}
\label{section:srplfg}
The above-mentioned features were calculated for every transaction. Because our prediction task focus is on relationships, we need to aggregate the information from all the transactions between a sender and a receiver. This aggregation is done in a slightly different way depending on the features used as shown in Table ~\ref{table:Aggregations}.

\begin{table}
	\caption{Aggregation type for features. Feature aggregation is performed for all transactions within a relationship.}
	\label{table:Aggregations}
	\centering
	\begin{tabular}{|p{1.7cm}|p{6.1cm}|}
		\hline
		\textbf{Aggregation} & \textbf{Features}\\
		\hline
		{Maximum} & {All the sentiment features}\\
		\hline
		{Minimum, \mbox{Maximum}, Median} & {Length of transactions, number of words, 
		the longest word length, proportion of word breaks to the message description length\tablefootnote{This feature helps to identify when spaces were removed}}\\
		\hline
		{Sum} & {All the toxicity features} \\
		\hline
		{Mean} & {All the emotion features, transaction amount, number of lower case words, number of upper case words, number of mixed case words, number of punctuation found} \\
		\hline
	\end{tabular}
	\end{table}

We also used additional features derived from all transactions in a relationship. These are the number of transactions sent in a relationship, the maximum number of transactions sent in a single day and the number of unique days in that a transactions has occured.

\subsection{Reciprocity}
\label{section:Reciprocity}
We also include the same features that are calculated on the replies of a relationship. That is, we calculate features on relationship  $\langle{a,b}\rangle$ as well as features on the reciprocal relationship  $\langle{b,a}\rangle$. 
This is to confirm our hypothesis that reciprocity might be useful as a recipient often avoids replying to an abusive sender. 
 
\section{Results and Discussion}
\label{section:results}
First, we performed an experiment to investigate what sets of features are able to discriminate the best between highly abusive and non-abusive cases. We evaluated the models with the following combinations of the feature sets: transaction details (TRX), simple text (ST), and toxicity and sentiment features (ETS). We show the results of repeated 5-fold cross-validation in Table~\ref{table:FeatureCombinations} and use precision, recall, F1,  AUC and AUC-PR metrics for evaluation. Overall, the best performing model was the random forest model using simple text, transaction details, emotion, toxicity and sentiment features and recipricity combined (ETS + ST + TRX). After selecting the best sets of features, we experimented with adding reciprocal features and observed further improvements (see Table~\ref{table:reciprocity})

\begin{table}
	\caption{Experimental results for combinations of emotion, toxicity and sentiment (ETS), simple text (ST) and transaction (TRX) features.}
	\label{table:FeatureCombinations}
	\centering
	\begin{tabular}{|p{1.5em}|p{1.5em}|p{1.5em}||p{2.3em}|p{2.3em}|p{2.3em}|p{2.3em}|p{2.3em}|}
		\hline
		\multicolumn{3}{|l||}{\textbf{Features}} &  \multicolumn{5}{|l|}{\textbf{Performance}}\\
		\hline
		\textbf{ETS} & \textbf{ST} & \textbf{TRX} & \textbf{Prec} & \textbf{Rec} & \textbf{F1} & \textbf{AUC-PR} & \textbf{ROC AUC}\\
		\hline
		{\checkmark} & {} & {} & {0.618} & {\textbf{0.740}} & {0.670} & {0.526} & {0.766} \\
		\hline
		{} & {\checkmark} & {} & {0.615} & {0.686} & {0.645} & {0.505} & {0.748} \\
		\hline
		{} & {} & {\checkmark} & {0.575} & {0.666} & {0.614} & {0.474} & {0.731} \\
		\hline
		{\checkmark} & {\checkmark} & {} & {0.633} & {0.728} & {0.671} & {0.531} & {0.775} \\
		\hline
		{} & {\checkmark} & {\checkmark} & {0.657} & {0.721} & {0.683} & {0.547} & {0.790} \\
		\hline
		{\checkmark} & {} & {\checkmark} & {0.638} & {0.709} & {0.669} & {0.532} & {0.780} \\
		\hline
		{\checkmark} & {\checkmark} & {\checkmark} & {\textbf{0.659}} & {0.730} & {\textbf{0.690}} & {\textbf{0.554}} & {\textbf{0.795}} \\
		\hline
	\end{tabular}
	\end{table}
	
\begin{figure}
	\centering
	\includegraphics[width=1\linewidth]{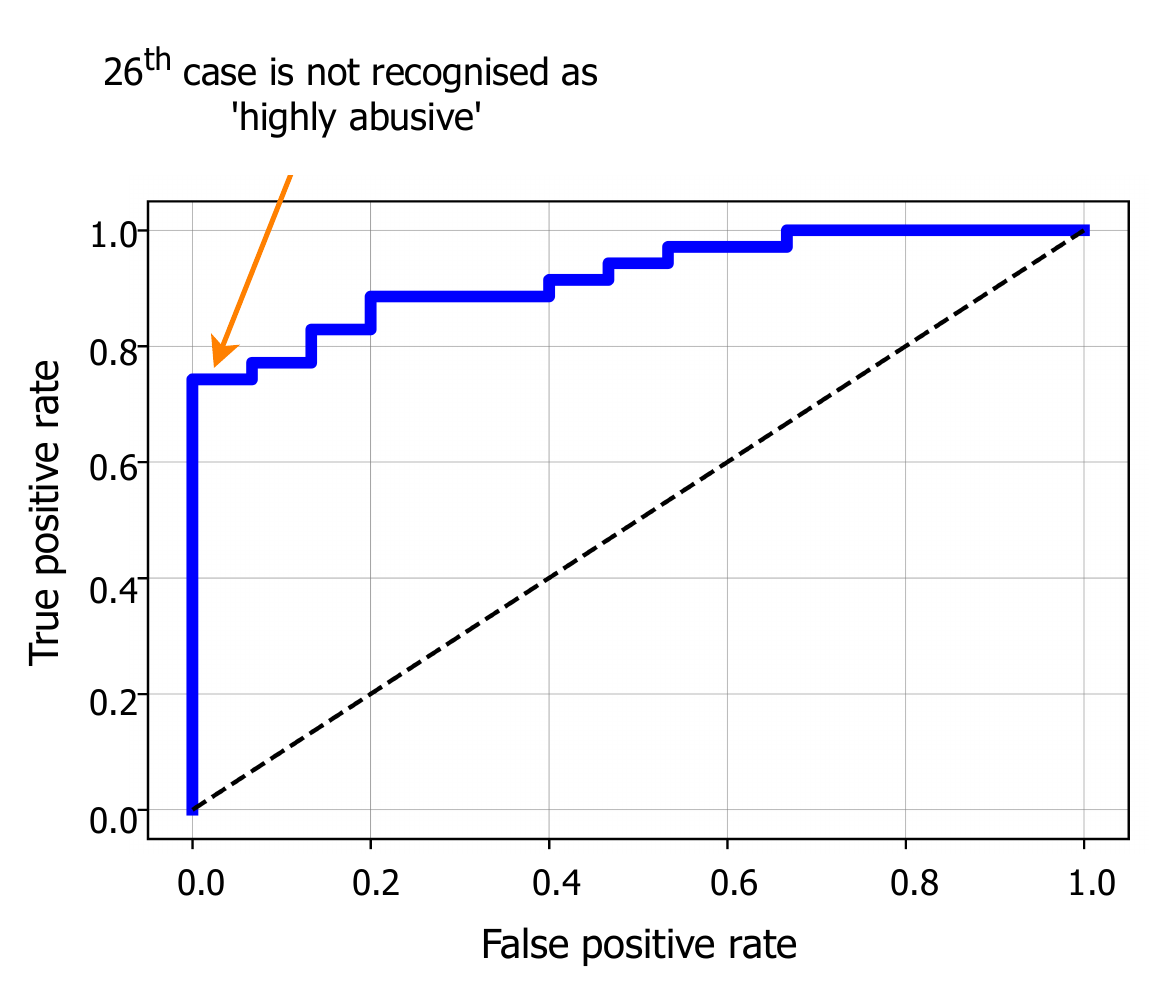}
	\caption{ROC curve (blue line) for the top 50 cases with the highest assigned probability of out-of-sample test set, which is predicted against our best model (ETS+ST+TRX) + reciprocity. Here the black dotted line is the ROC for a model with an AUC equal to 0.5 which is the perfectly diagonal line and it represents a model that makes random classifications.}
	\label{fig:roc_st_ets_trx_month_plus_2_month_forPaper_model1_forOutOfSet_50}	
\end{figure}

Next, we evaluated our results on an out-of-sample test set as outlined in Section~\ref{section:data}. The best system  (ETS + ST + TRX + reciprocity) of the previous experiment was used to demonstrate the capability of our model. The aim of this validation was to make sure that the model had consistency and had no false positives for the highest scored results, thus, allowing us to confidently select top cases for a manual review. Table~\ref{fig:roc_st_ets_trx_month_plus_2_month_forPaper_model1_forOutOfSet_50} displays the ROC curve of the models on the out-of-sample test set of the transaction descriptions collected over one month, whith no overlap of this month of data with our training dataset.

	
	\begin{table}
	\caption{Experimental results with the addition of reciprocity to the best model that contains emotion, toxicity and sentiment (ETS), simple text (ST) and transaction (TRX) features.}
	\label{table:reciprocity}
	\centering
	\begin{tabular}{|c|c|c|c|c|}

		\hline
		 \textbf{Prec} &\textbf{Rec} & \textbf{F1} & \textbf{AUC-PR} & \textbf{ROC AUC}\\
		\hline
	  {0.678} & {0.738} & {0.703} & {0.570} & {0.800}\\
		\hline
	\end{tabular}
	\end{table}
	
Next, the top 50 cases of sender-recipient pairs are manually labelled and used to produce the ROC curve, however, we did not manually verify the rest of the cases as it contains hundreds of transaction descriptions and requires a lot of  manual effort. The ROC curve shows the trade-off between sensitivity (true positive rate or recall), and specificity (1 - false positive rate). A black dotted line in Figure~\ref{fig:roc_st_ets_trx_month_plus_2_month_forPaper_model1_forOutOfSet_50} corresponds to a random guess. Note that classifiers that produce curves closer to the top-left corner indicate a better performance for the highest-scored cases.
From Fig. ~\ref{fig:roc_st_ets_trx_month_plus_2_month_forPaper_model1_forOutOfSet_50} we can clearly see that the best system predicts the highly abusive cases successfully with the first true negative in the 26th case. 

\section{Conclusion}
\label{section:conclusion}
In this paper we outlined a new problem related to the harassment and abuse happening in the financial services' domain. While resembling similarities with the other online social platforms, this problem poses new challenges and requires careful consideration. We outlined a particular case of abuse in transaction descriptions in the largest bank of Australia, and suggested ways to resolve it. We explored models with different feature sets and measured their performance in a real-life scenario. We showed that the best performing model is a supervised model trained on a variety of features that range in complexity from simple transaction and text features to the toxicity and emotions features that are calculated using state-of-the art advances in the field of NLP. The final model is already fully operational in the bank. To increase the model's robustness, we regularly retrain it when the sent cases are verified from the customer vulnarability specialists. 

We are continue improving our system in order to provide better protection to our customers. There are a range of potential improvements we are currently working on and aim to published in future work. Some examples of potential improvements are: better foreign language coverage, use of several months conversation history to detect a long-term abuse, use of BERT embeddings instead of high-level features when labelled training set is large enough.

\bibliographystyle{IEEEtran}
\bibliography{main}

\begin{thebibliography}{10}
\providecommand{\url}[1]{#1}
\csname url@samestyle\endcsname
\providecommand{\newblock}{\relax}
\providecommand{\bibinfo}[2]{#2}
\providecommand{\BIBentrySTDinterwordspacing}{\spaceskip=0pt\relax}
\providecommand{\BIBentryALTinterwordstretchfactor}{4}
\providecommand{\BIBentryALTinterwordspacing}{\spaceskip=\fontdimen2\font plus
\BIBentryALTinterwordstretchfactor\fontdimen3\font minus
  \fontdimen4\font\relax}
\providecommand{\BIBforeignlanguage}[2]{{%
\expandafter\ifx\csname l@#1\endcsname\relax
\typeout{** WARNING: IEEEtran.bst: No hyphenation pattern has been}%
\typeout{** loaded for the language `#1'. Using the pattern for}%
\typeout{** the default language instead.}%
\else
\language=\csname l@#1\endcsname
\fi
#2}}
\providecommand{\BIBdecl}{\relax}
\BIBdecl

\bibitem{social}
C.~Forlani, ``we are social: digital trends,''
  \url{https://wearesocial.com/au/blog/2021/10/social-media-users-pass-the-4-5-billion-mark/},
  2021.

\bibitem{dragiewicz2019domestic}
M.~Dragiewicz, B.~Harris, D.~Woodlock, M.~Salter, H.~Easton, A.~Lynch,
  H.~Campbell, J.~Leach, and L.~Milne, ``Domestic violence and communication
  technology: Survivor experiences of intrusion, surveillance, and identity
  crime,'' 2019.

\bibitem{fiolet2021exploring}
R.~Fiolet, C.~Brown, M.~Wellington, K.~Bentley, and K.~Hegarty, ``Exploring the
  impact of technology-facilitated abuse and its relationship with domestic
  violence: A qualitative study on experts’ perceptions,'' \emph{Global
  qualitative nursing research}, vol.~8, p. 23333936211028176, 2021.

\bibitem{flynn2021technology}
A.~Flynn, A.~Powell, and S.~Hindes, ``Technology-facilitated abuse: A survey of
  support services stakeholders,'' 2021.

\bibitem{austrac2021}
A.~for the Commonwealth~of Australia~2021, ``Austrac financial crime guide -
  preventing misuse and criminal communication through payment text fields,''
  https://www.austrac.gov.au/business/how-comply-guidance-and-resources/guidance-resources/payment-text-fields,
  2021.

\bibitem{reservebank2022payments}
E.~Connolly, ``Real-time payments in australia,'' in \emph{Opening Address to
  the Real Time Payments Summit 21/22}.\hskip 1em plus 0.5em minus 0.4em\relax
  Reserve Bank of Australia, 2022.

\bibitem{gabriel_news}
UOL, ``Homem é preso suspeito de fazer série de pix de centavos com ameaças
  à ex,''
  \url{https://noticias.uol.com.br/cotidiano/ultimas-noticias/2022/06/20/homem-e-preso-suspeito-de-fazer-pix-com-ameacas-a-ex.htm},
  2022.

\bibitem{majorbanks}
A.~Kane, ``Major banks reveal major issue of abusive transactions,''
  \url{https://www.theadviser.com.au/breaking-news/41494-major-banks-reveal-major-issue-of-abusive-transaction},
  2021.

\bibitem{humanloop}
\BIBentryALTinterwordspacing
H.~in~the Loop, ``What is a human in the loop?'' 2020. [Online]. Available:
  \url{https://humansintheloop.org/what-is-a-human-in-the-loop/}
\BIBentrySTDinterwordspacing

\bibitem{AIpolicy}
\BIBentryALTinterwordspacing
S.~Australian Government Department~of Industry and Resources, ``Australia's ai
  ethics principles,'' 2022. [Online]. Available:
  \url{https://www.industry.gov.au/publications/australias-artificial-intelligence-ethics-framework/australias-ai-ethics-principles}
\BIBentrySTDinterwordspacing

\bibitem{penzeymoog2021technology}
E.~PenzeyMoog and D.~C. Slakoff, ``As technology evolves, so does domestic
  violence: Modern-day tech abuse and possible solutions,'' in \emph{The
  Emerald International Handbook of Technology Facilitated Violence and
  Abuse}.\hskip 1em plus 0.5em minus 0.4em\relax Emerald Publishing Limited,
  2021.

\bibitem{zampieri2019semeval}
M.~Zampieri, S.~Malmasi, P.~Nakov, S.~Rosenthal, N.~Farra, and R.~Kumar,
  ``Semeval-2019 task 6: Identifying and categorizing offensive language in
  social media (offenseval),'' \emph{arXiv preprint arXiv:1903.08983}, 2019.

\bibitem{wang2021entailment}
S.~Wang, H.~Fang, M.~Khabsa, H.~Mao, and H.~Ma, ``Entailment as few-shot
  learner,'' \emph{arXiv preprint arXiv:2104.14690}, 2021.

\bibitem{chen2012detecting}
Y.~Chen, Y.~Zhou, S.~Zhu, and H.~Xu, ``Detecting offensive language in social
  media to protect adolescent online safety,'' in \emph{2012 International
  Conference on Privacy, Security, Risk and Trust and 2012 International
  Confernece on Social Computing}.\hskip 1em plus 0.5em minus 0.4em\relax IEEE,
  2012, pp. 71--80.

\bibitem{chatzakou2017measuring}
D.~Chatzakou, N.~Kourtellis, J.~Blackburn, E.~De~Cristofaro, G.~Stringhini, and
  A.~Vakali, ``Measuring\# gamergate: A tale of hate, sexism, and bullying,''
  in \emph{Proceedings of the 26th international conference on world wide web
  companion}, 2017, pp. 1285--1290.

\bibitem{raj2021cyberbullying}
C.~Raj, A.~Agarwal, G.~Bharathy, B.~Narayan, and M.~Prasad, ``Cyberbullying
  detection: Hybrid models based on machine learning and natural language
  processing techniques,'' \emph{Electronics}, vol.~10, p. 2810, 2021.

\bibitem{devlin2018bert}
J.~Devlin, M.-W. Chang, K.~Lee, and K.~Toutanova, ``Bert: Pre-training of deep
  bidirectional transformers for language understanding,'' \emph{arXiv preprint
  arXiv:1810.04805}, 2018.

\bibitem{liu2019roberta}
Y.~Liu, M.~Ott, N.~Goyal, J.~Du, M.~Joshi, D.~Chen, O.~Levy, M.~Lewis,
  L.~Zettlemoyer, and V.~Stoyanov, ``Roberta: A robustly optimized bert
  pretraining approach,'' \emph{arXiv preprint arXiv:1907.11692}, 2019.

\bibitem{bodapati2019neural}
S.~B. Bodapati, S.~Gella, K.~Bhattacharjee, and Y.~Al-Onaizan, ``Neural word
  decomposition models for abusive language detection,'' \emph{arXiv preprint
  arXiv:1910.01043}, 2019.

\bibitem{azzouza2019twitterbert}
N.~Azzouza, K.~Akli-Astouati, and R.~Ibrahim, ``Twitterbert: Framework for
  twitter sentiment analysis based on pre-trained language model
  representations,'' in \emph{International Conference of Reliable Information
  and Communication Technology}.\hskip 1em plus 0.5em minus 0.4em\relax
  Springer, 2019, pp. 428--437.

\bibitem{csiro2020cyberbullying}
D.~Andreas, C.~Cacron, R.~M, B.~S, C.~N, M.~J, W.~V, and J.~B. Wan~S, Paris~C,
  ``Cyberbullying detection from social media in the australian context,''
  \emph{Report EP203561, CSIRO Australia}, 2020.

\bibitem{ribeiro2017like}
M.~H. Ribeiro, P.~H. Calais, Y.~A. Santos, V.~A. Almeida, and W.~Meira~Jr, ``"
  like sheep among wolves": Characterizing hateful users on twitter,''
  \emph{arXiv preprint arXiv:1801.00317}, 2017.

\bibitem{detoxify}
L.~Hanu, ``Detoxify,'' \url{https://github.com/unitaryai/detoxify}, 2020.

\bibitem{hugging_emotions}
\BIBentryALTinterwordspacing
B.~Savani, ``Distilbert-base-uncased-emotion model,'' 2020. [Online].
  Available:
  \url{https://huggingface.co/bhadresh-savani/distilbert-base-uncased-emotion}
\BIBentrySTDinterwordspacing

\bibitem{hutto2014vader}
C.~Hutto and E.~Gilbert, ``Vader: A parsimonious rule-based model for sentiment
  analysis of social media text,'' in \emph{Proceedings of the International
  AAAI Conference on Web and Social Media}, vol.~8, 2014.

\end{thebibliography}


 

%



\end{document}